%% file: language_effect_swebench.tex
\documentclass[11pt,twocolumn]{article}
\usepackage[top=1in,bottom=1in,left=1in,right=1in]{geometry}
\usepackage{times}
\usepackage{graphicx}
\usepackage{float}
\usepackage{amsmath,amssymb}
\usepackage{natbib}
\usepackage[hyphens]{url}
\usepackage{hyperref}
\usepackage{booktabs}
\usepackage{multirow}
\usepackage{siunitx}
\usepackage[utf8]{inputenc}

\hypersetup{
    colorlinks=true,
    linkcolor=blue,
    filecolor=magenta,
    urlcolor=cyan,
}

\title{Mythbuster: Chinese Language Is Not More Efficient Than English in Vibe Coding: \\
A Preliminary Study on Token Cost and Problem-Solving Rate}
\author{Simiao Ren$^\dagger$\footnote{Correspondence: benren@scam.ai}, Xingyu Shen$^*$, Yuchen Zhou$^*$, Dennis (Tsang) Ng, Ankit Raj\\$^{*, \dagger}$Scam.ai\\$^*$Equal contribution\\$^\dagger$Corresponding author}
\date{}

\begin{document}

\maketitle

\begin{figure*}[!htbp]
\centering
\includegraphics[width=0.8\textwidth]{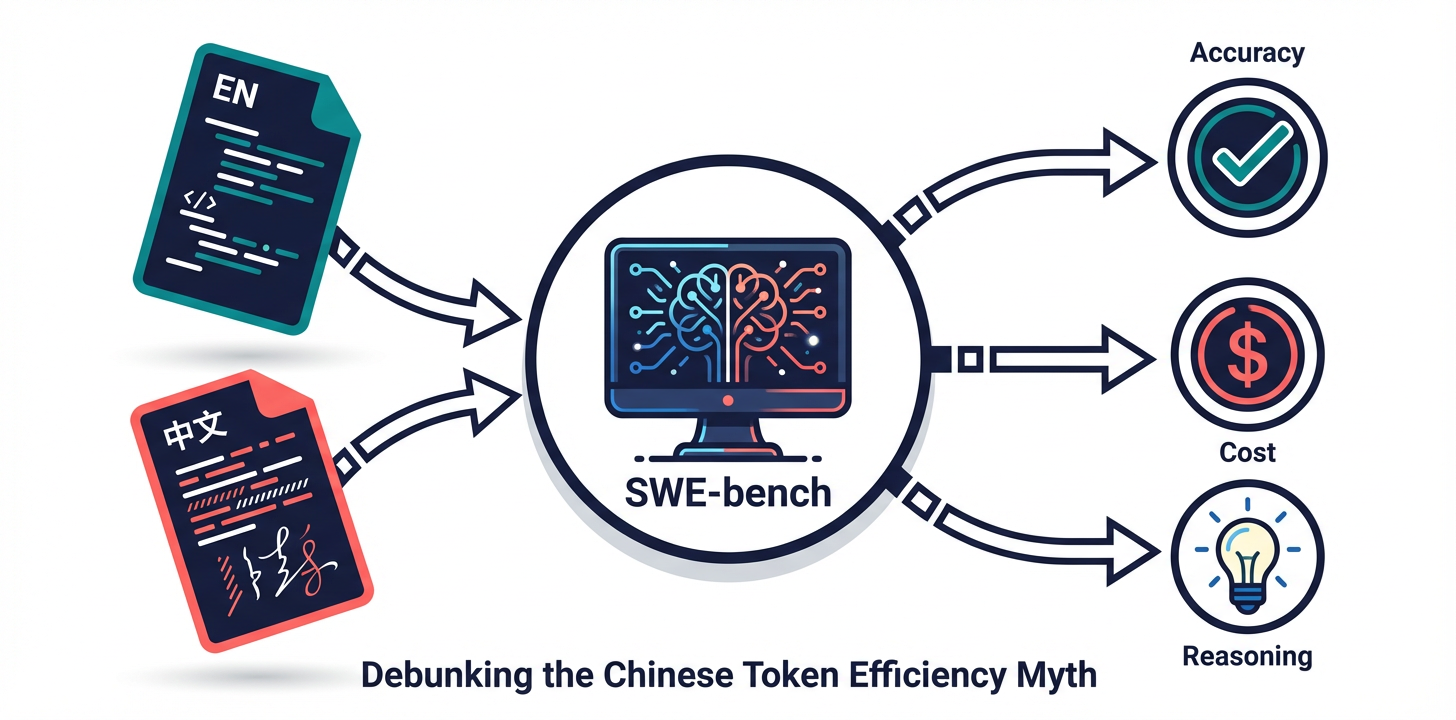}
\caption{Graphical abstract showing the research workflow and key findings.}
\label{fig:abstract}
\end{figure*}

\begin{abstract}
A claim has been circulating on social media and practitioner forums that Chinese prompts are more token-efficient than English for LLM coding tasks, potentially reducing costs by up to 40\%. This claim has influenced developers to consider switching to Chinese for ``vibe coding'' to save on API costs. In this paper, we conduct a rigorous empirical study using SWE-bench Lite, a benchmark of software engineering tasks, to evaluate whether this claim of Chinese token efficiency holds up to scrutiny. Our results reveal three key findings: First, the efficiency advantage of Chinese is not observed. Second, token cost varies by model architecture in ways that defy simple assumptions: while MiniMax-2.7 shows 1.28x higher token costs for Chinese, GLM-5 actually consumes fewer tokens with Chinese prompts. Third, and most importantly, we found that the success rate when prompting in Chinese is generally lower than in English across all models we tested. We also measure cost efficiency as expected cost per successful task---jointly accounting for token consumption and task resolution rate. These findings should be interpreted as preliminary evidence rather than a definitive conclusion, given the limited number of models evaluated and the narrow set of benchmarks tested due to resource constraints; they indicate that language effects on token cost are model-dependent, and that practitioners should not expect cost savings or performance gains just by switching their prompt language to Chinese.

\end{abstract}

\section{Introduction}

The deployment of large language models (LLMs) for software engineering tasks has gained significant traction, with models such as GPT-4, Claude, and open-source alternatives demonstrating remarkable capabilities in code generation, bug detection, and automated repair~\citep{feng2020swebench,chen2021humaneval}. AI-assisted coding has grown explosively: 84\% of developers now use or plan to use AI coding tools~\citep{stackoverflow2025survey}, and AI is estimated to generate approximately 29\% of new U.S. software code as of late 2024~\citep{tamm2026aicoding}. The ``vibe coding'' paradigm, where developers use natural language to generate code through LLMs, has become mainstream, with platforms reaching \$100M ARR in as little as eight months~\citep{lovable2025arr} and the AI coding tools market projected to reach \$37B by 2032~\citep{snsinsider2025market}. As token-based API pricing dominates LLM economics, with output tokens typically priced 4--5$\times$ higher than input tokens, even modest efficiency gains translate to substantial cost savings at scale.

A claim has been circulating on social media and practitioner forums that Chinese prompts are more token-efficient than English for coding tasks, potentially reducing API costs by up to 40\%~\citep{YouTubeShort}. This claim has influenced developers to consider switching to Chinese for ``vibe coding'' to save on token costs.

\begin{figure}[!htbp]
\centering
\includegraphics[width=0.35\textwidth]{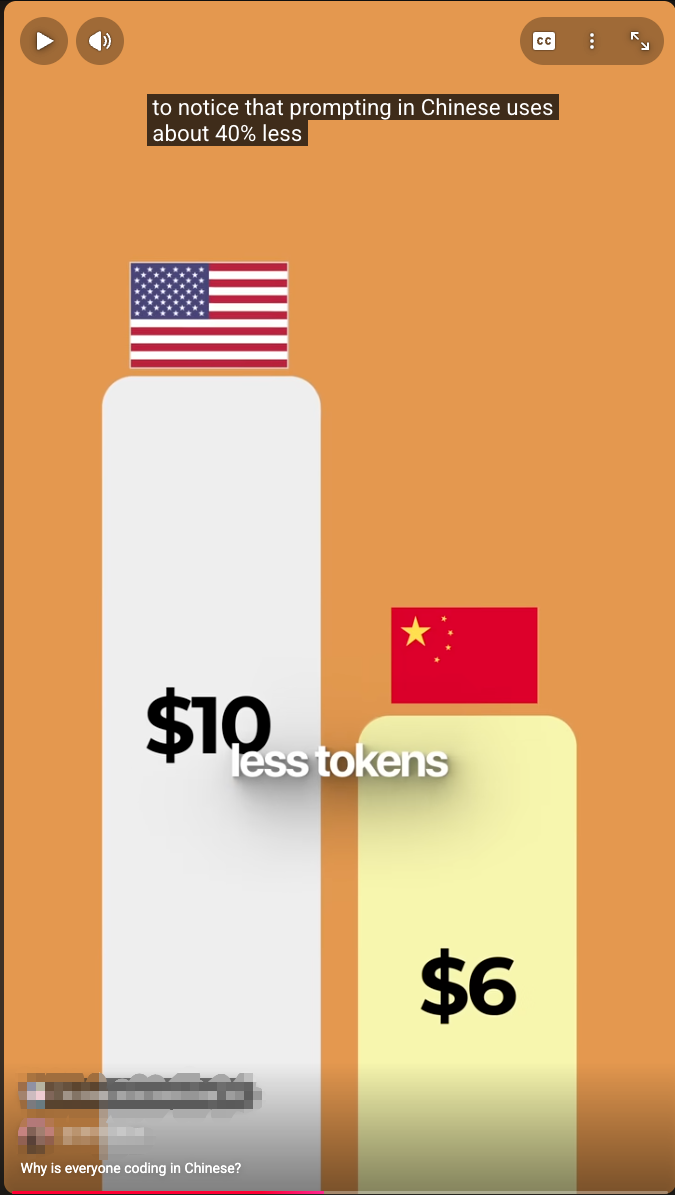}
\caption{The popular social media claim that Chinese prompts are 40\% more token-efficient, motivating this empirical study. This YouTube short has garnered over 2.6 million views and 63k likes, with top comments endorsing the claim. Notably, the original post does not cite any empirical research to support this assertion.}
\label{fig:motivation}
\end{figure}

The social media claim typically asserts that Chinese is inherently more efficient because Chinese characters are more information-dense than English text. However, this claim conflates three distinct phenomena: (1) information density---the fact that Chinese characters can express ideas with fewer characters at the semantic level; (2) tokenizer allocation---tokenizers optimized for English training data may fragment CJK characters into multiple tokens; and (3) reasoning verbosity---some anecdotal evidence suggests Chinese chain-of-thought reasoning produces shorter output text. Without rigorous empirical validation, the popular efficiency narrative remains an unverified claim that could misguide developers.

Recent academic research has begun to clarify these issues. Research on tokenization premiums demonstrates that different languages require different numbers of tokens to express equivalent content depending on tokenizer vocabulary allocation, with English often having the lowest token premium for GPT-style tokenizers while Chinese may require substantially more tokens per character~\citep{petrov2023unfairness}. Similarly, recent model technical reports describe dedicating expanded tokenizer vocabulary to non-English languages to improve compression, acknowledging that multilingual tokenization efficiency is a deliberate design choice~\citep{qwen2tech2024,cui2023chinesellama}.

Despite this emerging understanding, rigorous empirical validation on domain-specific tasks like code debugging remains scarce. Most prior studies have compared English performance against machine-translated Chinese prompts without controlling for confounding factors such as model architecture and token limits. The few multilingual evaluations that exist often focus on general reasoning benchmarks rather than software engineering tasks, where precise technical understanding is critical and where token costs can accumulate substantially over long agentic interactions.

In this paper, we conduct a controlled empirical study to test the popular claim that Chinese prompts are more token-efficient than English for code reasoning tasks. Our research question is: \textit{Does the choice of prompt language (English vs Chinese) significantly affect token cost and problem-solving rate in LLM-based code reasoning, and if so, which language is actually more efficient?} We focus specifically on SWE-bench Lite, selecting 50 instances from its 300 real-world software engineering tasks, each requiring models to understand bug descriptions, locate relevant code, and generate patches that pass test suites.

Our contributions are threefold. First, we provide a systematic comparison of English and Chinese prompts across three LLM families (MiniMax-2.7, GPT-5.4-mini, and GLM-5) on identical task instances. Second, we measure token consumption at the API level, distinguishing between prompt, completion, and reasoning tokens. Third, we analyze whether observed performance differences can be attributed to language per se or to confounding factors such as tokenization efficiency.

\section{Related Work}

\subsection{Multilingual LLM Performance}

Prior work on multilingual LLM evaluation has established that model capability varies significantly across languages, with English typically serving as the best-supported language due to its dominance in training data~\citep{pires2019multilingual,zhang2023m3exam}. Studies on Chinese LLM performance have documented higher token-to-character ratios compared to English, due to the sub-word tokenization challenges inherent in logographic writing systems~\citep{petrov2023unfairness,sennrich2015subword,kudo2018sentencepiece}. However, these studies have largely focused on general language understanding tasks rather than code-specific reasoning.

Recent research on tokenization premiums demonstrates that token efficiency is tokenizer-dependent: English often has the lowest token premium for GPT-style tokenizers, while Chinese typically requires more tokens per character~\citep{petrov2023unfairness}. This work contradicts the simplistic narrative that Chinese is inherently token-efficient; instead, efficiency depends on how tokenizer vocabulary was allocated during training. Similarly, recent large language model technical reports explicitly expand tokenizer vocabulary dedicated to non-English languages to address this gap, acknowledging that multilingual compression requires deliberate design investment~\citep{qwen2tech2024,cui2023chinesellama}.

\subsection{Code Generation Benchmarks}

SWE-bench represents a significant advancement in evaluating LLMs on real-world software engineering tasks~\citep{feng2020swebench}. Unlike synthetic benchmarks such as HumanEval~\citep{chen2021humaneval}, SWE-bench provides authentic bug reports from popular open-source projects, requiring models to understand natural language descriptions, locate relevant source files, and generate precise patches.

Multilingual code generation has been studied on benchmarks such as HumanEval-XL~\citep{humanevalxl2024}, which documents that prompt language influences accuracy and pass@k rates. However, these benchmarks rarely report token consumption, focusing instead on accuracy metrics. This gap in reporting makes it difficult to assess whether observed accuracy differences correlate with token efficiency.

\subsection{Token Efficiency in LLMs}

Token efficiency, broadly defined as the ratio of useful output to total token consumption, has become increasingly important as API pricing for LLMs can constitute significant operational costs. Prior work has documented that prompt length correlates with task difficulty and that longer contexts can increase token consumption substantially.

Practitioner discussions on platforms such as Reddit and practitioner blogs have produced anecdotal evidence about Chinese token efficiency. Token counting experiments using GPT-4o's tokenizer found that Chinese text maps to substantially fewer characters per token (approximately 1.33 chars/token) compared to English (approximately 4.75 chars/token)~\citep{castillo2025tokens}, but this does not imply lower total token usage when accounting for meaning density and model behavior. These informal findings highlight the need for controlled experiments that measure total token consumption, not just input compression ratios.

\section{Benchmark: SWE-bench}

We employ SWE-bench Lite, a curated subset of 300 software engineering tasks drawn from popular Python open-source projects including Django, matplotlib, pytest, and scikit-learn~\citep{feng2020swebench}. Due to the substantial token costs associated with running agentic code generation at scale (each instance requiring thousands of API calls with 1,500 iteration limits), we selected 50 instances for this study. This subset was chosen using stratified random sampling (seed=42) to ensure representation across repository categories while remaining within our computational budget. Each task presents a bug report containing a natural language description of the issue, along with a failing test case that must pass after the patch is applied.

SWE-bench has emerged as a \emph{de facto} standard for evaluating LLMs on real-world software engineering tasks. Major AI labs now routinely report SWE-bench performance in their model launches: OpenAI reported 33.2\% on SWE-bench Verified for GPT-4o~\citep{openai2024swebench}, while Anthropic reported 49\% for Claude 3.5 Sonnet~\citep{anthropic2025swebench}, noting substantial improvements over prior state-of-the-art. This widespread adoption validates SWE-bench as a meaningful proxy for practical coding capability.

SWE-bench provides a rigorous evaluation framework because it requires not merely syntactic correctness but semantic correctness---the generated patch must actually resolve the reported bug as verified by test execution. This stands in contrast to simpler metrics such as patch application success, which can be achieved by patches that do not properly address the underlying issue.

The benchmark presents several advantages for studying language effects. First, the task descriptions are relatively standardized in structure (problem description plus test case), reducing variability in prompt complexity. Second, the evaluation is deterministic: a patch either passes all tests or fails, eliminating subjective judgment. Third, the benchmark has been widely adopted, providing a well-understood and challenging evaluation context for software engineering tasks.

\section{Methods}

\subsection{Experimental Design}

We conducted a two-phase evaluation for each model-language combination. In Phase 1 (patch generation), we used the MiniSWEAgent framework to interact with each SWE-bench instance. The agent received the task description in either English or Chinese and was allocated a step limit of 1,500 iterations with a 2-hour time budget. In Phase 2 (evaluation), we applied the generated patches to the original codebase using the official SWE-bench evaluation harness and executed the test suite to determine resolution status.

We evaluated three model families: MiniMax-2.7 (via OpenRouter), GPT-5.4-mini (OpenAI via OpenRouter), and GLM-5 (Z.ai via OpenRouter). All models were accessed through the OpenRouter API to ensure consistent evaluation infrastructure and centralized token tracking.

\textbf{Model Selection Rationale.} We selected these three models to span a range of architectures and tokenization strategies. MiniMax-2.7 and GLM-5 are Chinese-developed models with tokenizer vocabularies that may include CJK-specific optimizations; GPT-5.4-mini is primarily trained on English-dominant data and serves as a contrast. This selection allows us to test the hypothesis that Chinese-developed models may handle Chinese prompts more efficiently, while also measuring the language effect across diverse architectures.

\subsection{Prompt Translation}

The English task descriptions were professionally translated into Chinese by a bilingual software engineer with domain expertise. We preserved the technical terminology and code-related phrasing to ensure semantic equivalence. The Chinese prompts retained the same structure as the English versions (PR description format), differing only in language.

\subsection{Task Distribution}

We selected 50 instances from SWE-bench Lite using stratified random sampling (seed=42) to ensure representation across repository categories. Our sample includes instances from 12 repositories: Django (22), SymPy (6), matplotlib (5), scikit-learn (5), pytest-dev (3), sphinx-doc (3), astropy (1), psf/requests (1), pylint-dev (1), pydata/xarray (1), mwaskom/seaborn (1), and pallets/flask (1). This distribution mirrors the original SWE-bench Lite composition, where Django dominates due to its popularity in open-source bug reports.

\subsection{Token Tracking}

We collected token usage data from the OpenRouter activity logs, which record per-API-call metrics including prompt tokens, completion tokens, reasoning tokens (where applicable), and cost. This granular tracking enables us to analyze token efficiency at the instance level and aggregate across experimental conditions.

\section{Results}

\subsection{Resolution Rates}

\begin{figure}[!h]
\centering
\includegraphics[width=0.48\textwidth]{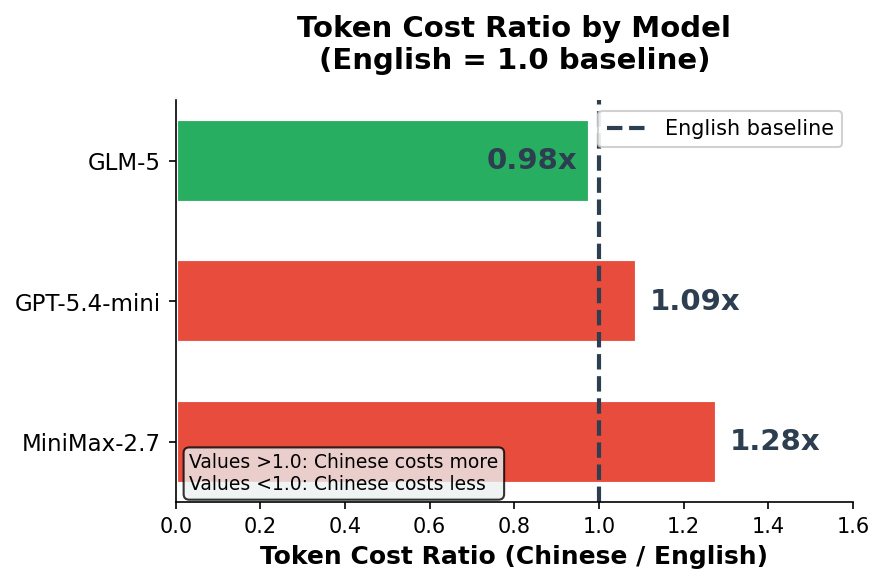}
\caption{Comprehensive results visualization showing (A) resolution rates by language, and (B) normalized token cost ratio (Chinese/English) where English serves as the 1.0 baseline per model. Values above 1.0 indicate Chinese costs more tokens; values below 1.0 indicate Chinese costs fewer tokens.}
\label{fig:results}
\end{figure}

Figure~\ref{fig:results} presents the comprehensive results visualization, and Table~\ref{tab:results} provides the detailed metrics. The most striking finding is the minimal language effect for the best-performing model: MiniMax-2.7 achieves 66.0\% resolution on English and 61.5\% on Chinese, a difference of only 4.5 percentage points. Given the small sample size of 50 instances, this difference is likely within the range of statistical noise, though we do not perform formal significance testing.

\textbf{Instance Count Note.} The number of evaluable instances varies across conditions (EN: 50, ZH: 39--49). This variation arises from API errors during patch generation where some instances produced empty or truncated patches, likely due to token limit pressure from the longer Chinese prompts. These instances are excluded from the resolution rate calculation. Notably, MiniMax-2.7 ZH dropped to 39 evaluable instances (22\% fewer than EN), a non-trivial gap that may introduce upward bias in ZH resolution rates if the excluded instances were systematically harder; we discuss this in the Limitations section.

For comparison, GPT-5.4-mini shows a larger gap (9.9 percentage points: 36.0\% English vs 26.1\% Chinese), and GLM-5 shows a similar gap (9.5 percentage points: 64.6\% English vs 55.1\% Chinese). However, these differences are still modest compared to the 30 percentage point gap between the best and worst models.

\subsection{Token Cost Analysis}

\begin{figure*}[!htbp]
\centering
\includegraphics[width=0.9\textwidth]{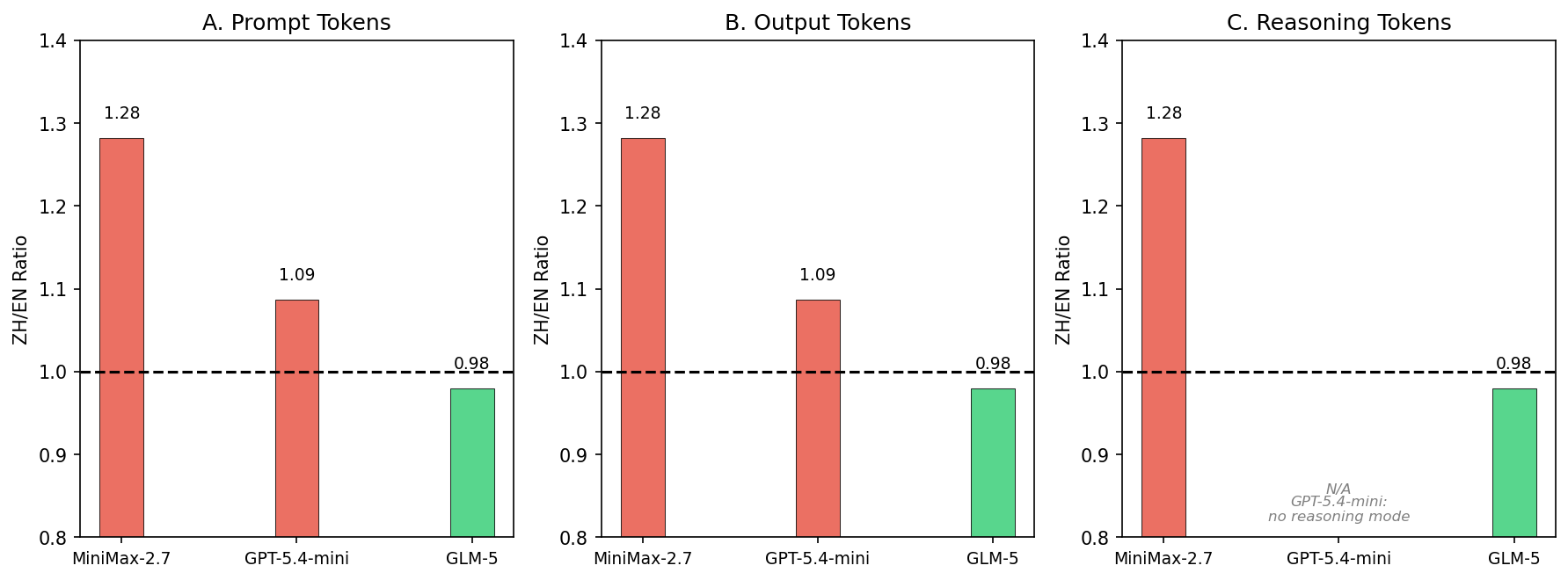}
\label{fig:token_breakdown}
\caption{ZH/EN token ratio by model and token type (English = 1.0 baseline). Values above 1.0 indicate Chinese requires more tokens; values below 1.0 indicate Chinese requires fewer tokens. GPT-5.4-mini has no reasoning mode.}
\end{figure*}

The token cost analysis reveals important nuances (Table~\ref{tab:results} and Figure~\ref{fig:token_breakdown}). Contrary to the common assumption that Chinese prompts are universally more token-expensive, we observe model-dependent behavior. MiniMax-2.7 shows the expected pattern: Chinese prompts require 1.28x more tokens than English (382,974 vs 298,720 prompt tokens per instance). GPT-5.4-mini shows a more modest 1.09x ratio (91,681 vs 84,347). Critically, GLM-5 actually consumes \emph{fewer} tokens for Chinese prompts (1,388,094 vs 1,417,012), a 0.98x ratio.

This finding challenges the simple narrative that Chinese prompts are inherently less token-efficient. Rather, the tokenization behavior depends on how each model's tokenizer was trained and whether it has been optimized for CJK character sets.

\textbf{Expected Cost per Successful Task.} While we report raw token consumption, token counts alone do not fully capture cost efficiency in practical deployments. A more appropriate measure is the \emph{expected cost per successful task}, which depends jointly on token usage and task resolution rate:

\begin{equation}
C_{\text{eff}} = \frac{\text{Avg.\ Token Cost per Attempt}}{\text{Resolution Rate}}
\label{eq:cost_efficiency}
\end{equation}

This formulation captures a critical but often overlooked factor: the cost of failed attempts. Even if a language produces fewer tokens per attempt, a lower task success rate leads to higher overall cost because failed attempts must be retried or abandoned. Consider GPT-5.4-mini: although Chinese prompts incur only a modest 1.09x token overhead relative to English, the resolution rate drops 9.9 percentage points (from 36.0\% to 26.1\%). Applying Equation~\ref{eq:cost_efficiency}, this resolution penalty substantially amplifies the effective cost of Chinese prompts, making the language switch counterproductive even from a pure cost perspective. The raw token ratio thus paints a misleadingly optimistic picture of Chinese efficiency; the correct cost-efficiency lens must incorporate task success probability.

Applying Equation~\ref{eq:cost_efficiency} across all three evaluated models reveals a consistent pattern in our data: we do not observe a systematic cost-efficiency gain from Chinese prompts in any model-language combination once resolution rate is incorporated. For MiniMax-2.7, both token costs and resolution rates move against Chinese (1.28x higher tokens, 4.5pp lower resolution rate), producing a compounded penalty. For GPT-5.4-mini, the token overhead is modest (1.09x) but the resolution penalty (9.9pp) dominates. Only GLM-5 shows marginal token savings for Chinese (0.98x), yet these are partially offset by its 9.5pp resolution rate decline.

\textbf{Cost Decomposition.} The effective cost of a language-model combination can be decomposed into three interacting factors: (1) \emph{prompt token length}, determined by tokenizer vocabulary and language compactness; (2) \emph{reasoning and output verbosity}, reflecting how the model generates chain-of-thought and code in a given language; and (3) \emph{retry frequency induced by failure}, which multiplies the per-attempt cost by the expected number of attempts before success. Our results suggest that retry frequency is the dominant amplifying factor: resolution rate differences of 4.5--9.9 percentage points translate to substantially higher expected total costs than the raw token ratios (0.98--1.28x) would imply in isolation. This decomposition highlights why cost analyses that focus solely on input token compression---the metric typically cited in social media comparisons---systematically underestimate the true cost impact of language choice.

\subsection{Reasoning Tokens}

The three models differ substantially in their use of explicit reasoning tokens (Table~\ref{tab:results}). GLM-5 exhibits the highest reasoning token count (approximately 87,000 per instance), consistent with its explicit chain-of-thought mode. MiniMax-2.7 shows moderate reasoning token usage (22,000-29,000 per instance). GPT-5.4-mini shows zero explicit reasoning tokens, as this model does not support or utilize a reasoning chain mode. In our data, higher reasoning token usage is associated with higher resolution rates, suggesting that explicit step-by-step reasoning may contribute to better code understanding and patch generation; however, given only three model data points, a definitive causal claim cannot be made.

\section{Discussion}

Our results carry several implications for the deployment of LLMs in multilingual software engineering contexts.

\subsection{The Claim of Chinese Efficiency is Not Supported}

The popular claim that Chinese prompts are more token-efficient than English is not supported by our empirical data. For the best-performing model (MiniMax-2.7), the 4.5 percentage point difference between English and Chinese suggests that any efficiency advantage of Chinese is negligible at best. More importantly, we found that Chinese prompts do not consistently reduce token costs: MiniMax-2.7 actually requires \emph{more} tokens for Chinese (1.28x), while GLM-5 requires \emph{fewer} tokens for Chinese (0.98x).

The gap between the best and worst models (approximately 30 percentage points in resolution rate) far exceeds any language effect observed, reinforcing that model selection is the primary determinant of performance. Language effects on token cost are themselves \emph{mediated by architecture}: the effect depends entirely on the interplay between the model's tokenizer vocabulary, training data distribution, and reasoning mode. A practitioner cannot predict whether Chinese will be more or less token-efficient without knowing which model they are using. Users who switched to Chinese based on social media claims should not expect meaningful cost savings.

The widespread belief that Chinese prompts might reduce token costs conflates three distinct phenomena that are often discussed interchangeably but have different implications.

\textbf{Information density} refers to the fact that Chinese characters often express ideas with fewer characters than their English equivalents at the semantic level. However, tokenizers do not operate on characters directly; they operate on sub-word units, meaning character-level efficiency does not automatically translate to token efficiency.

\textbf{Tokenizer allocation} refers to how tokenizer vocabulary is designed and trained. Research on tokenization premiums shows that English often has the lowest premium on GPT-style tokenizers because English dominates training data~\citep{petrov2023unfairness}. Chinese characters are frequently fragmented into multiple tokens (e.g., the character for ``cat'' may map to 3 tokens under some tokenizers). This actually explains why Chinese prompts often require \emph{more} tokens than English, contrary to the social media claim.

\textbf{Reasoning verbosity} refers to observed differences in output text length when models reason in different languages. Some social media experiments have reported that Chinese chain-of-thought reasoning produces shorter text than English, which would reduce output tokens. However, this concerns output tokens, not the total token cost of a task, and may not hold for code generation where precision is required.

Critically, the total token cost of a task depends on the sum of input and output tokens, modulated by success probability. A language that produces shorter reasoning but fails more often may actually increase total token cost due to retries. Our study accounts for this by measuring per-instance token consumption across successful and unsuccessful attempts, providing a more complete picture than input-token-only analyses.

\subsection{Empirical Validation of Tokenizer Allocation Effects}

To directly test whether tokenizer vocabulary design mediates the relationship between language and token efficiency, we measured token counts across five representative tokenizers using 23 SWE-bench Lite problem descriptions in both English and Chinese (see Appendix Table~\ref{tab:tokenizer_comparison} for detailed results). The tokenizers span a spectrum from English-dominant (cl100k\_base, used by GPT and Llama) to Chinese-native (GLM's chatglm3-6b tokenizer) to CJK-expanded (Qwen2-7B).

The results confirm that tokenizer vocabulary allocation is the decisive factor, not language properties per se. The GLM tokenizer, trained with substantial Chinese data, produces \emph{fewer} tokens for Chinese than English (ZH/EN ratio of 0.923), directly inverting the typical pattern. In contrast, GPT/Llama's cl100k\_base tokenizer, optimized for English, penalizes Chinese with 15\% more tokens per equivalent content. Qwen2 occupies an intermediate position with only 2.5\% overhead, reflecting its expanded CJK vocabulary~\citep{qwen2tech2024}. These findings align with prior work documenting substantial tokenization premiums for non-English languages under English-dominant vocabularies~\citep{petrov2023unfairness}.

Notably, Chinese text is \emph{always} more character-dense across all tokenizers (1.96--2.72 chars/token) compared to English (2.69--3.72 chars/token). This confirms that the "Chinese is efficient" narrative conflates two distinct phenomena: character-level information density (where Chinese characters do encode more semantic information per glyph) versus tokenizer-level compression (where English-optimized vocabularies fragment CJK characters into multiple sub-word units). The former is a property of the writing system; the latter is a design choice in tokenizer training. Our empirical data demonstrates that the tokenizer choice determines which language appears more token-efficient, and that there is no intrinsic efficiency advantage to Chinese that transfers across tokenizers.

\begin{figure}[!h]
\centering
\includegraphics[width=0.48\textwidth]{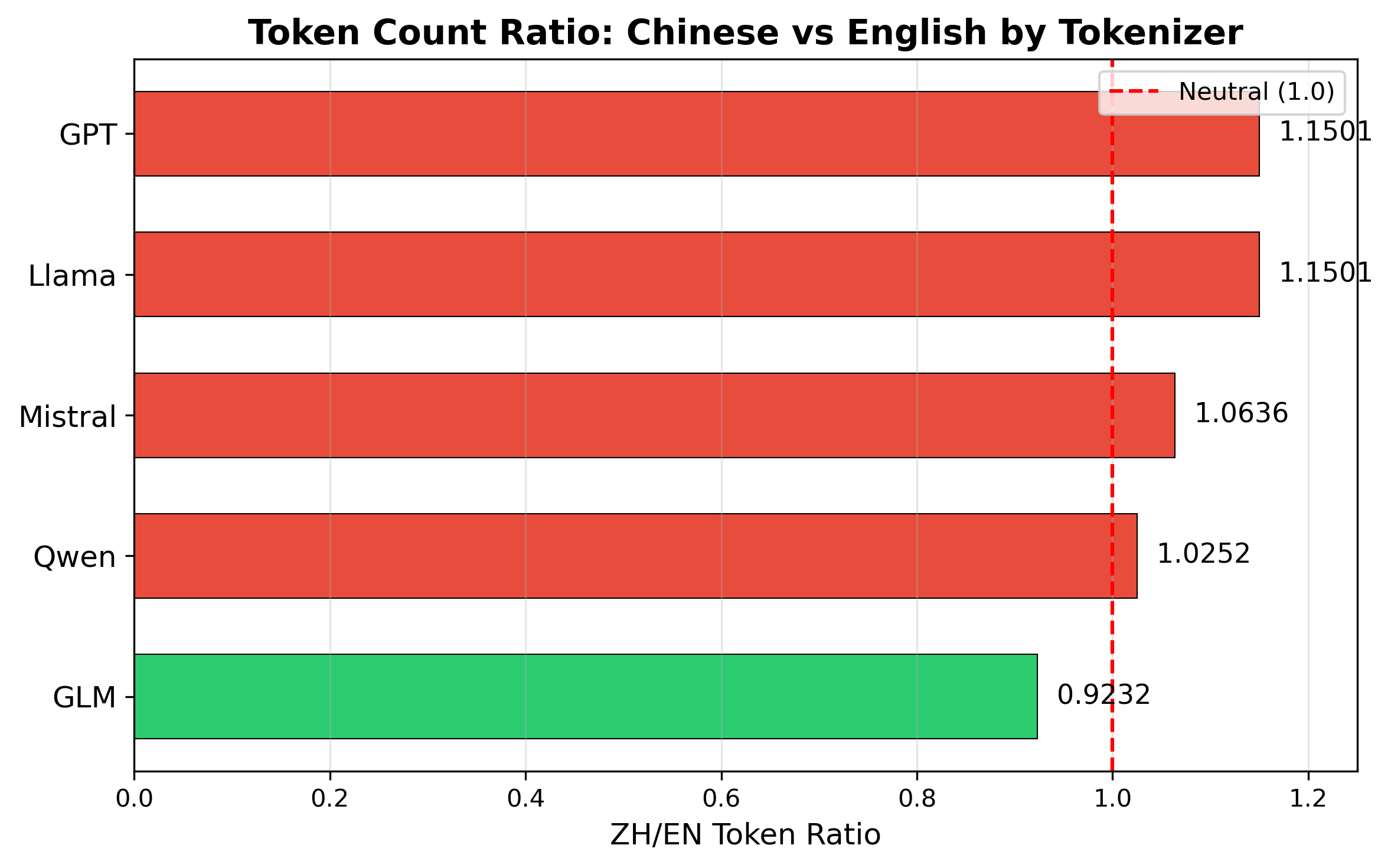}
\caption{Token count ratio (Chinese/English) across different tokenizer families. Values above 1.0 indicate Chinese requires more tokens; values below 1.0 indicate Chinese requires fewer tokens.}
\label{fig:tokenizer_ratio}
\end{figure}

\begin{figure}[!h]
\centering
\includegraphics[width=0.48\textwidth]{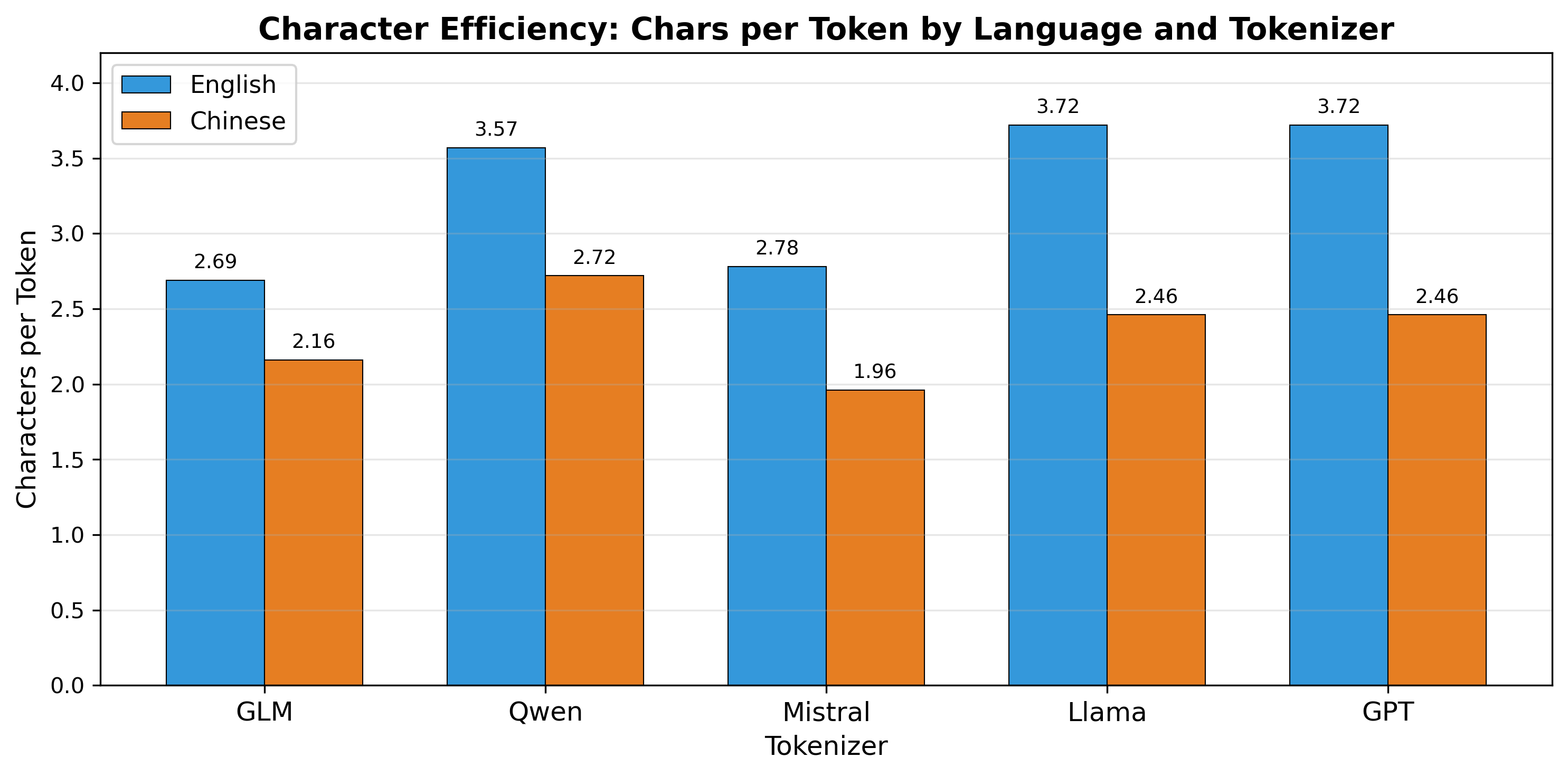}
\caption{Character efficiency (chars/token) for English and Chinese across different tokenizer families. Higher values indicate more characters encoded per token, reflecting greater compression efficiency.}
\label{fig:chars_per_token}
\end{figure}

\subsection{Practical Implications}

For practitioners deploying LLMs for coding tasks, our findings carry direct implications. First, switching to Chinese for ``vibe coding'' to reduce API costs is unlikely to yield the claimed 30--40\% savings, and may increase effective costs once the resolution rate penalty is taken into account. Second, model selection yields far greater cost-performance gains than language choice: the resolution rate gap between MiniMax-2.7 and GPT-5.4-mini is approximately 30 percentage points---an order of magnitude larger than any observed language effect. Third, for users working primarily in Chinese, selecting a model with tokenizer vocabulary aligned to CJK text (such as GLM-5 or MiniMax-2.7) is likely to yield greater token efficiency than prompting in a mismatched language. In practice, optimizing for cost efficiency should prioritize model selection and task success probability, rather than prompt language, which acts as a second-order factor whose sign and magnitude depend entirely on the model in use. Practitioners optimizing API costs should prioritize model choice and task batching over language switching.

\subsection{Limitations}

This study represents a preliminary investigation into multilingual code reasoning efficiency rather than a comprehensive benchmark evaluation, and our findings should be interpreted as such. Several limitations apply. First, we evaluate only 50 SWE-bench instances; larger-scale studies would provide more statistical power to detect small language effects. Additionally, the number of evaluable ZH instances is lower than EN for MiniMax-2.7 (39 vs.\ 50), because instances that produced empty patches due to token limit pressure were excluded. If excluded instances were systematically more difficult, reported ZH resolution rates may be slightly inflated relative to EN; future work should enforce equal instance coverage across conditions. Second, our model selection was constrained by API availability and cost considerations. Rather than aiming for broad model coverage, our selection intentionally captures qualitatively different tokenizer regimes: an English-dominant tokenizer without a reasoning chain mode (GPT-5.4-mini), a CJK-optimized tokenizer with moderate reasoning token usage (MiniMax-2.7), and a CJK model with a dedicated chain-of-thought reasoning mode producing substantially higher reasoning token counts (GLM-5). This design allows us to identify how language effects vary across distinct architectural axes---tokenizer vocabulary allocation and reasoning mode---rather than simply accumulating data points within a single design family. Nevertheless, broader and more definitive conclusions require evaluation across additional model families including GPT-4o, Claude, Llama, and Mistral, and we caution against over-generalizing from three models. Third, we used a fixed step limit of 1,500 iterations, which may disadvantage models with different reasoning strategies. Fourth, we study only English and Chinese; other languages (Korean, Japanese, Arabic) may exhibit qualitatively different tokenization and reasoning behavior.

\textbf{Token Counting Methodology.} A critical caveat concerns how different API providers report token usage. MiniMax and potentially other models report \emph{cumulative} prompt tokens, where each API call includes the full conversation context accumulated to that point. In contrast, GPT-5.4-mini reports \emph{per-call} prompt tokens for only that specific request. This means raw prompt token counts are not directly comparable across providers. For cost comparisons, we rely on the API-reported cost per instance, which is provider-independent and represents the actual economic cost of each model-language combination. Future studies should implement unified token tracking at the API level to enable precise per-token efficiency comparisons.

\section{Conclusion}

We conducted a rigorous empirical study to test the popular social media claim that Chinese prompts are more token-efficient than English for LLM-based code reasoning on SWE-bench. Our key finding is straightforward: the myth is busted. Chinese prompts are not more efficient for vibe coding, at least as of today. Specifically:

\begin{itemize}
    \item Language effects are small (4.5--9.9 percentage points in resolution rate) and model-dependent: Chinese is more token-efficient on some models (GLM-5: 0.98x) but less efficient on others (MiniMax-2.7: 1.28x).
    \item Model selection dominates over language choice: the 30 percentage point gap between models far exceeds any language effect.
    \item The belief that Chinese is intrinsically more token-efficient conflates character-level density with tokenizer-level compression; tokenizer design, not language properties, determines efficiency.
\end{itemize}

Practitioners should focus on model selection rather than prompt language for cost-performance optimization.

\appendix
\section*{Appendix: Supporting Tables}

\subsection*{Prompt Comparison}

\input{tables/prompt_comparison}

\subsection*{Resolution Rates and Token Metrics}

\input{tables/results_table}

\subsection*{Tokenizer Efficiency Comparison}

\begin{table}[!h]
\centering
\caption{ZH/EN token ratios and character efficiency across five tokenizers on 23 SWE-bench Lite instances.}
\label{tab:tokenizer_comparison}
\begin{tabular}{lccc}
\toprule
Tokenizer & ZH/EN Ratio & EN chars/tok & ZH chars/tok \\
\midrule
GLM (chatglm3-6b) & 0.923 & 2.69 & 2.16 \\
Qwen (Qwen2-7B) & 1.025 & 3.57 & 2.72 \\
Mistral (Mistral-7B) & 1.064 & 2.78 & 1.96 \\
Llama (cl100k\_base) & 1.150 & 3.72 & 2.46 \\
GPT (cl100k\_base) & 1.150 & 3.72 & 2.46 \\
\bottomrule
\end{tabular}
\end{table}

\emergencystretch=1em
\bibliographystyle{plain}
\bibliography{references}

\end{document}

%% file: tables/prompt_comparison.tex
\begin{table*}[htbp]
\centering
\caption{Side-by-Side Comparison of English and Chinese Prompts for the Same SWE-bench Instance (sqlfluff\_\_sqlfluff-1625)}
\label{tab:prompt_comparison}
\begin{tabular}{p{0.47\linewidth}p{0.47\linewidth}}
\toprule
\textbf{English Prompt} & \textbf{Chinese Prompt (original characters)} \\
\midrule
TSQL - L031 incorrectly triggers ``Avoid using aliases in join condition'' when no join present

\textbf{Expected Behaviour}

Both queries should pass; only difference is the addition of a table alias `a':

\begin{verbatim}
SELECT [hello]
FROM
    mytable
\end{verbatim}

and with alias:

\begin{verbatim}
SELECT a.[hello]
FROM
    mytable AS a
\end{verbatim}

\textbf{Observed Behaviour}

1/ passes
2/ fails with: L031: Avoid using aliases in join condition.

No join condition present :-)

\textbf{Steps to Reproduce}

Lint queries above

\textbf{Dialect:} TSQL
\textbf{Version:} sqlfluff 0.6.9, Python 3.6.9
 &
\textbf{Chinese Prompt Translation}

\medskip
\includegraphics[width=\linewidth]{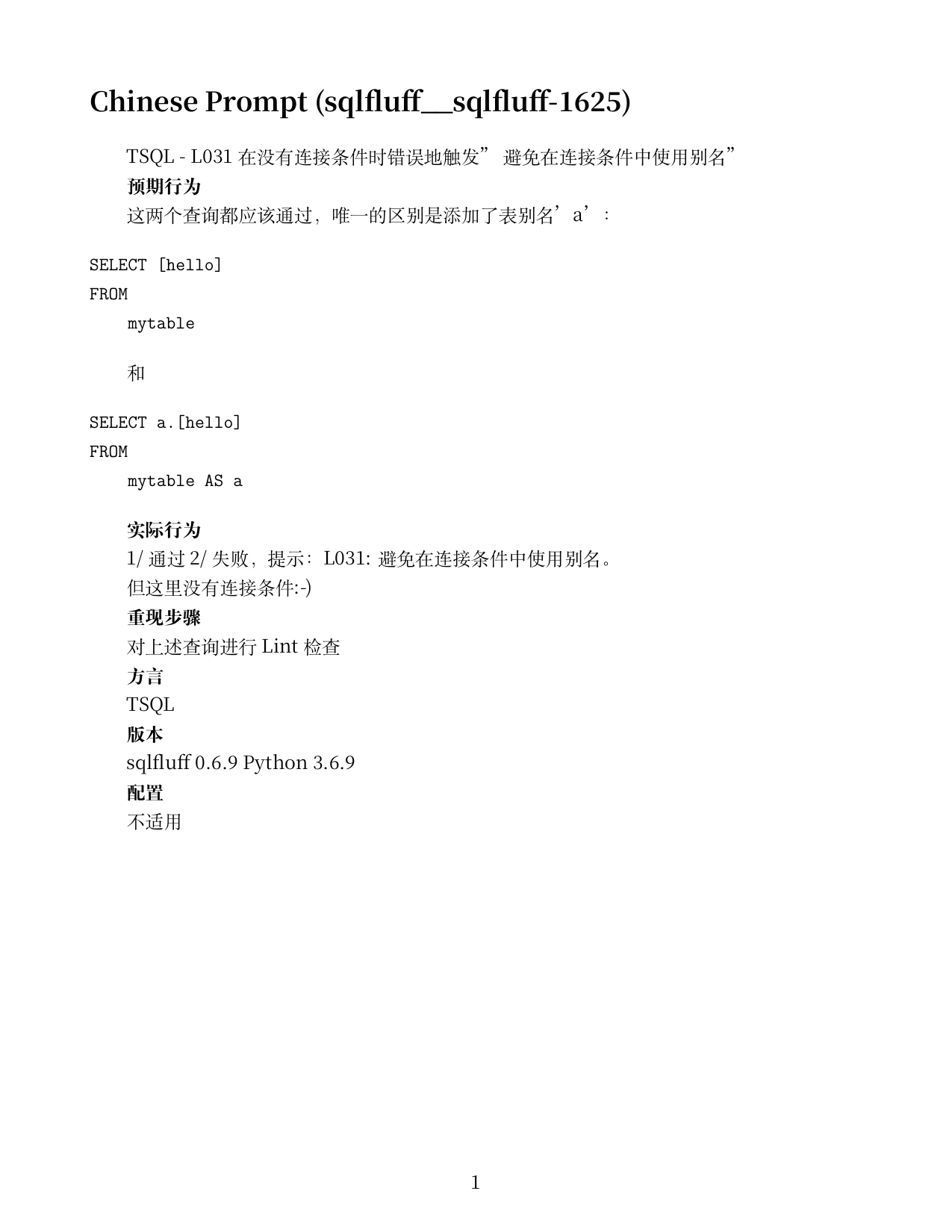}
\\
\bottomrule
\end{tabular}
\par\smallskip
\noindent\textit{Note:} The Chinese prompt is a professional translation preserving technical terminology and structure. Both prompts produce identical task outcomes but differ in token count due to language-specific tokenization.
\end{table*}

%% file: tables/results_table.tex
\begin{table*}[htbp]
\centering
\caption{Resolution Rates and Token Metrics by Model and Language}
\label{tab:results}
\begin{tabular}{@{}lccccccc@{}}
\toprule
\textbf{Model} & \textbf{Lang} & \textbf{Inst} & \textbf{Res} & \textbf{Rate} & \textbf{Prompt/Inst} & \textbf{Output/Inst} & \textbf{Reason/Inst} \\
\midrule
MiniMax-2.7 & EN & 50 & 33 & 66.0\% & 298,720 & 61,762 & 22,368 \\
MiniMax-2.7 & ZH & 39 & 24 & 61.5\% & 382,974 & 79,182 & 28,677 \\
GPT-5.4-mini & EN & 50 & 18 & 36.0\% & 84,347 & 2,695 & 0\textsuperscript{a} \\
GPT-5.4-mini & ZH & 46 & 12 & 26.1\% & 91,681 & 2,929 & 0\textsuperscript{a} \\
GLM-5 & EN & 48 & 31 & 64.6\% & 1,417,012 & 112,803 & 87,474 \\
GLM-5 & ZH & 49 & 27 & 55.1\% & 1,388,094 & 110,501 & 85,688 \\
\bottomrule
\end{tabular}
\par\smallskip
\noindent\textit{Note:} \textsuperscript{a}GPT-5.4-mini does not support explicit reasoning chains; reasoning tokens are zero. Instance counts below 50 indicate runs that errored out or did not complete successfully.
\end{table*}